\documentclass{article}

\usepackage{microtype}
\usepackage{graphicx}
\usepackage{subfigure}
\usepackage{booktabs} 
\usepackage{subcaption}
\usepackage{amsmath,amssymb,amsthm}
\newtheorem{theorem}{Theorem}
\usepackage{xurl}
\usepackage{booktabs}
\usepackage[table]{xcolor}
\definecolor{headergray}{HTML}{E6E6E6}
\definecolor{bestcyan}{HTML}{C5EEF0}
\newcommand{\bestcell}[1]{\cellcolor{bestcyan}\textbf{#1}}
\usepackage{multirow}
\usepackage{float}
\usepackage{pifont}
\usepackage[hyperfootnotes=false]{hyperref}

\usepackage[accepted]{mlsys2025}

\mlsystitlerunning{PrismServe: Efficient LLM Serving on Heterogeneous Accelerator Clusters}

\begin{document}

\twocolumn[
\mlsystitle{PrismServe: Efficient LLM Serving on Heterogeneous Accelerator Clusters}




\begin{mlsysauthorlist}
\mlsysauthor{Jinhao Zhang}{bupt}
\mlsysauthor{Zicheng Yan}{ustc}
\mlsysauthor{Boyang Zhang}{ict}
\mlsysauthor{Zeyu Liu}{ict}
\mlsysauthor{Daning Cheng}{ict}
\mlsysauthor{Yunquan Zhang}{ict}
\mlsysauthor{Guangming Tan}{ict}
\mlsysauthor{Xiang Gao}{zjlab}
\end{mlsysauthorlist}

\mlsysaffiliation{bupt}{Beijing University of Posts and Telecommunications, Beijing, China}
\mlsysaffiliation{ustc}{University of Science and Technology of China, Hefei, China}
\mlsysaffiliation{ict}{Institute of Computing Technology, Chinese Academy of Sciences, Beijing, China}
\mlsysaffiliation{zjlab}{Zhejiang Lab, Hangzhou, China}

\mlsyscorrespondingauthor{Daning Cheng}{DaningCheng@ict.ac.cn}

\mlsyskeywords{Machine Learning, MLSys}

\vskip 0.3in

\begin{abstract}
Serving large language models across heterogeneous accelerators is challenging because partitioning a replica across device types introduces communication overhead during autoregressive decoding, while deploying complete quantized replicas can degrade model quality. We present PrismServe, a serving system that combines request-level collaboration across heterogeneous hardware with a mixture of quantized experts derived from the same foundation model. Each expert executes within a group of same-type accelerators in a single node, while a two-level router selects an expert based on predicted response quality and then selects a replica according to load and prefix-cache locality. Implemented on unmodified vLLM, PrismServe achieves 2.0--2.6$\times$ the throughput of cross-node or cross-vendor full-precision pipelined deployments and reduces P99 time per output token by 55--65\% when serving Qwen3-14B on 48 accelerators from three vendors. Across Qwen3-8B, Qwen3-14B, and Qwen3-32B on five downstream tasks, PrismServe outperforms the best single quantization method by 2.0--4.3 points while remaining within 1.4 points of full-precision accuracy. Our code is available at \url{https://anonymous.4open.science/r/LLM-Inference-System-B000/}.
\end{abstract}
]



\printAffiliationsAndNotice{} 


\section{Introduction}
\label{sec:intro0}
 
Large language models (LLMs) keep growing, and the demand for serving them is growing even faster. Open-weight models now exceed a trillion parameters~\citep{DBLP:journals/corr/abs-2407-21783,DBLP:journals/corr/abs-2412-19437,DBLP:journals/corr/abs-2606-19348}, and Google reported in May 2026 that its products process more than 3.2 quadrillion tokens per month, up from 9.7 trillion two years earlier~\citep{google2026io}. Meanwhile, AI accelerators span an increasingly diverse range of vendors and generations, each with its own software stack and collective-communication library. Few operators can acquire enough accelerators of a single type to keep pace with demand. Our own cluster illustrates this challenge: it comprises seven accelerator types from six vendors, with only a limited number of devices of each type. Using accelerators of different types together to serve the same model, which we call \emph{heterogeneous collaborative inference}, is therefore becoming a practical necessity.

Heterogeneous collaborative inference, however, runs into a communication wall. The standard way to serve a model on many accelerators is to partition it with tensor or pipeline parallelism~\citep{shoeybi2019megatron,huang2019gpipe}, which exchanges activations at every decoding step. Accelerators of one type do so over a vendor interconnect and collective-communication library. Accelerators of different types share neither: their libraries are mutually incompatible, cross-vendor collective libraries have only begun to appear, and every cross-type transfer must leave the vendor interconnect and pass through host memory or the datacenter network. A replica partitioned across types therefore crosses a slow link on every generated token, and its pipeline advances at the pace of the slowest vendor. On eight accelerators of our cluster, replicas pipelined across two vendors reach only 41--43\% of the throughput of the fastest single-node deployment at eight concurrent requests, and even pipelining across two nodes of the same vendor lowers throughput by 24--56\%.
 
We therefore take the opposite approach: accelerators of different types never communicate. Each request is served from start to finish by one group of same-type accelerators within a node, which communicate only over their own interconnect. Each group must then hold a complete model. For large models or memory-limited accelerators, a full-precision replica needs a group of several accelerators or even several nodes, which brings the communication back and leaves few groups to serve requests in parallel. Quantization only has to shrink the model enough for a small group of same-type accelerators within one node to hold it; in our deployment, each group consists of two accelerators. Existing systems sit on either side of this choice. Heterogeneity-aware serving systems optimize how a model is partitioned and placed across devices~\citep{jiang2023hexgen,jiang2025hexgen,mei2025helix,DBLP:conf/ppopp/ZhaoWWPL24}, but a partitioned replica still communicates on every token. Serving stacks that route requests across vendor-local replicas avoid this communication but deploy the same model everywhere, and a quantized model pays in accuracy: even the best single quantization method for each task trails full precision by 2.7--4.9 points (Table~\ref{tab:model_quality}).
 
This accuracy loss need not be accepted. Different quantization methods applied to the same model err on different inputs, so a pool of differently quantized variants can answer correctly where any single variant fails (\S\ref{sec:expert-complementarity}). PrismServe therefore stacks differently quantized variants of one model as \emph{quantized experts}, each hosted on its own group of same-type accelerators, and a lightweight quality router, trained to predict from the request alone which expert yields the lowest loss on the reference answer, sends each request to one expert. Stacking more experts raises accuracy further: adding an expert never increases the lowest loss achievable on any input, and in our evaluation, growing the pool from two to five experts improves accuracy by 1.5--2.1 points (\S\ref{sec:expert-pool-size}). Any quantization method can supply an expert, so each accelerator type can host whichever methods its own toolchain supports; we evaluate experts from AWQ~\citep{DBLP:conf/mlsys/0002TTYCWXDG024}, GPTQ~\citep{DBLP:journals/corr/abs-2210-17323}, FlatQuant~\citep{DBLP:conf/icml/SunLBBZLHY0Y0LY25}, QuaRot~\citep{DBLP:conf/nips/AshkboosMCLCJAH24}, and the 8-bit SmoothQuant~\citep{DBLP:conf/icml/XiaoLSWDH23}. Once dispatched, a request's prefill and decoding stay within one group. The groups together form a mixture of quantized experts hosted on heterogeneous hardware: the differences between accelerator types no longer lie on the decoding path, and the differences between quantization methods recover most of the accuracy that any single method loses.
 
We build PrismServe on unmodified vLLM~\citep{DBLP:conf/sosp/KwonLZ0ZY0ZS23} and the vLLM Router~\citep{vllmrouter}. A unified gateway runs a two-level routing pipeline (\S\ref{sec:inference_system}). The first level, a quality router built from a 0.6B-parameter embedding model~\citep{zhang2025qwen3} and a two-layer MLP, binds each request to one quantized expert before prefill, so its KV cache never migrates. The second level, a per-expert replica router, balances load and prefix-cache locality across that expert's replicas, which can run on any accelerator type whose runtime supports the expert's format. Because expert selection is decoupled from replica selection, operators can change the number of replicas per expert, or add nodes of another accelerator type, without retraining the quality router. Adding a quantization method leaves the inference engines and the existing experts unchanged; only the lightweight router is retrained.
 
PrismServe thus delivers accuracy and speed together. On 48 accelerators of our cluster, two nodes from each of three vendors (Ascend 910C, MTT S5000, and BI-V150), it serves Qwen3-14B~\citep{DBLP:journals/corr/abs-2505-09388} with 2.0--2.6$\times$ the throughput of full-precision replicas pipelined across nodes or vendors and cuts the P99 time per output token (TPOT) by 55--65\% (\S\ref{sec:performance-48cards}). Even against full-precision replicas confined to a single eight-accelerator node, it delivers 1.8--19.9\% higher throughput and 12.1--26.8\% lower P99 TPOT at eight concurrent requests (\S\ref{sec:deployment-8cards}). Across Qwen3-8B, 14B, and 32B, PrismServe stays within 1.4 points of full-precision accuracy on five downstream tasks and outperforms the best single quantization method on every task by 2.0--4.3 points.
 
In summary, this paper makes the following contributions:

\ding{182} A design for heterogeneous collaborative inference without cross-type communication, in which same-type accelerator groups sized to hold a quantized model host quantized experts from any quantization method, and a quality router trained on lowest-loss expert labels assigns each request to one expert, so that stacking more experts raises accuracy.

\ding{183} PrismServe, an implementation of this design on accelerators from three vendors that combines a gateway, a two-level router, and per-vendor vLLM backends without modifying the inference engine (\S\ref{sec:inference_system}).

\ding{184} An evaluation that measures, with Qwen3-14B on up to 48 accelerators, the cost of pipelining replicas across nodes and vendors and shows that PrismServe delivers 2.0--2.6$\times$ the throughput of cross-node full-precision serving, and that, on Qwen3-8B, 14B, and 32B, it stays within 1.4 points of full-precision accuracy (\S\ref{sec:serving-performance} and \S\ref{sec:model-quality}).

\section{Background and Motivation}
\label{sec:background}

The inference deployment of large-scale LLMs is jointly constrained by model weight storage, device memory capacity, and cross-device execution efficiency. Quantization reduces storage and memory requirements by lowering numerical precision, increasing deployment flexibility, but it may also degrade response quality. PrismServe focuses on two interrelated questions: how to exploit the complementarity of different quantization methods to improve model performance, and how to organize these quantized experts into an inference system suited to heterogeneous resources. This section explains PrismServe's motivation from the perspectives of input-dependent quality differences and cross-node execution constraints.

\subsection{Bottlenecks in Distributed Inference}
\label{sec:deployment-bottlenecks}
\begin{figure*}[t]
    \centering
    \includegraphics[width=.8\textwidth]{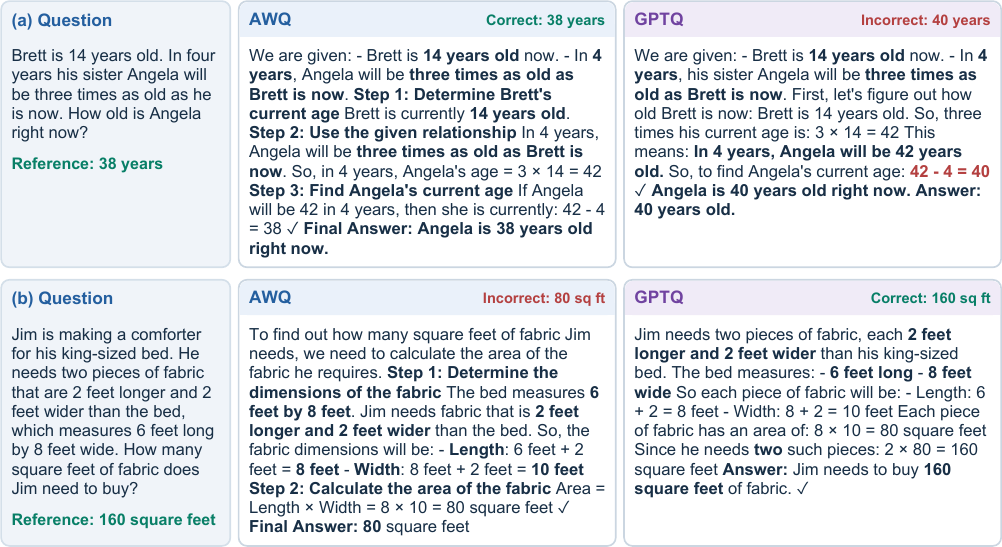}
    \caption{Complementary reasoning errors under AWQ and GPTQ quantization. (a) AWQ correctly solves the age problem, while GPTQ reaches an incorrect answer due to a subtraction error. (b) GPTQ correctly computes the total fabric area, while AWQ overlooks the requirement for two pieces. These examples illustrate that neither method consistently preserves correctness across problems.}
    \label{fig:case-study}
\end{figure*}
As LLMs grow in scale, the memory required for model weights, the KV cache, and runtime state also increases. When these requirements exceed a single node's GPU memory capacity, tensor parallelism (TP)~\cite{shoeybi2019megatron} and pipeline parallelism (PP)~\cite{huang2019gpipe} distribute the model across multiple GPUs and nodes. Insufficient homogeneous resources may further require combining devices from different vendors.

Cross-node deployment introduces communication into each request's execution path: TP exchanges and synchronizes intermediate results, while PP transfers activations between stages on different nodes. These operations recur for every generated token during autoregressive decoding. When communication cannot be effectively overlapped with computation, network transfers and synchronization waits increase inference latency. Cross-vendor deployment may additionally require host-mediated transfers because of incompatible communication backends, introducing further copying and coordination overhead.

Quantization reduces weight memory requirements, potentially allowing a complete model and the runtime state required by the target workload to fit within a single node or a homogeneous GPU group. The system can therefore deploy independent replicas on compatible hardware and distribute requests across heterogeneous resources, avoiding activation transfers between cross-vendor model stages within each request. PrismServe builds on this deployment opportunity by routing requests among complete quantized experts.

\begin{figure}[t]
    \centering
    \includegraphics[width=.92\linewidth]{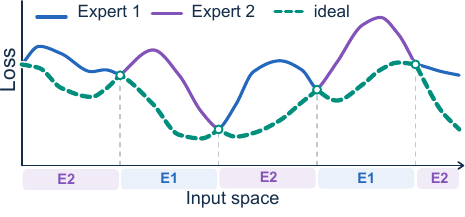}
    \caption{Conceptual illustration of input-dependent expert complementarity. The experts have different loss profiles across the input space. The dashed curve sketches an idealized routing objective, and the bottom labels indicate the preferred expert in each region.}
    \label{fig:complementarity}
\end{figure}

\subsection{Complementarity of Quantized Experts}
\label{sec:expert-complementarity}

The performance loss introduced by quantization is not uniformly distributed across inputs. Prior work~\citep{DBLP:conf/icml/LiuMZZYWLD25} studies this uneven impact across data groups as \emph{quantization unfairness}: performance gaps can widen even when overall accuracy is largely preserved. This finding motivates examining whether different quantization methods applied to the same LLM preserve quality on different inputs. If their errors do not fully overlap, input-dependent expert selection offers an opportunity to recover some of the quality lost when using a single quantized model.

Figure~\ref{fig:case-study} illustrates complementary errors between AWQ and GPTQ on two reasoning problems. In the age problem, AWQ produces the correct answer, whereas GPTQ makes an arithmetic error in the final step. In the fabric-area problem, GPTQ correctly accounts for the two required pieces, whereas AWQ overlooks this condition. Either fixed expert answers only one of these two problems correctly, while appropriate request-level selection can exploit the strengths of both. These examples illustrate why choosing a quantized variant solely by its aggregate task score can overlook another variant's advantages on individual inputs.

Figure~\ref{fig:complementarity} conceptually illustrates this opportunity: different experts achieve lower prediction loss in different regions of the input space, and the dashed curve sketches an idealized routing objective that exploits these differences. Online requests, however, do not provide reference answers for directly comparing expert losses. Executing all experts before selecting a response would also introduce additional inference overhead. The resulting challenge is to predict a suitable expert from the request input alone, exploiting complementarity while executing only one expert per request.

Together, these observations motivate exploiting expert complementarity to recover quality while retaining the deployment benefits of quantization. The goal is to combine efficient serving on heterogeneous hardware with better quality than a fixed quantized expert.

\section{Methodology}
PrismServe organizes variants of the same base model produced by different quantization methods into an expert pool, deploys them across heterogeneous resources, and coordinates inference through request-level routing. The system comprises two stages: offline training and online inference. During offline training, expert prediction losses on training samples provide supervision for an input-dependent expert-selection router. During online inference, the router selects an expert solely from the request content and dispatches the request to an instance hosting the full expert model on compatible hardware.

\subsection{Offline Training}
\label{sec:offline-training}
\begin{figure}[t]
    \centering
    \includegraphics[width=\linewidth]{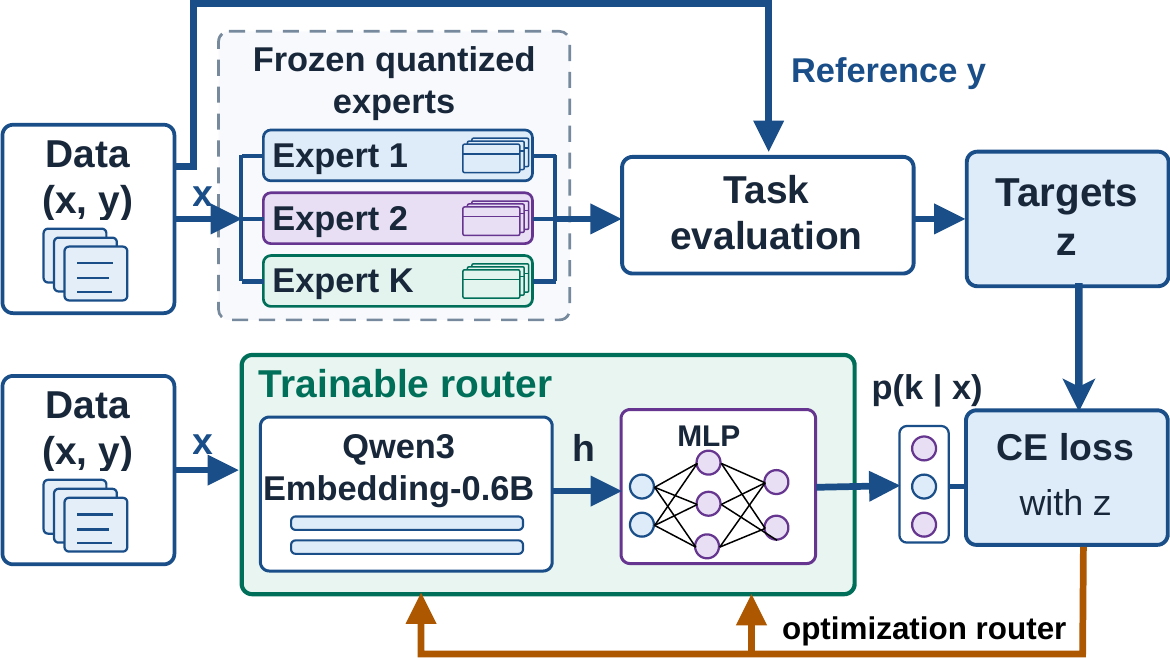}
    \caption{Offline training of the PrismServe router. Routing labels are determined by the expert with the lowest prediction loss on each training sample. Quantized experts remain frozen, while the embedding encoder and MLP are jointly trained using cross-entropy loss.}
    \label{fig:router_training}
\end{figure}

We construct the router training dataset from problem-solving, conversation, and document samples, with data sources and proportions detailed in appendix. A separate router is trained for each base model: Qwen3-8B, Qwen3-14B, and Qwen3-32B. Routing labels are constructed using the quantized experts associated with that base model. When the number or composition of experts changes, we reconstruct the labels within the current candidate set and retrain the corresponding router.

As shown in Figure~\ref{fig:router_training}, offline training consists of two steps. We first evaluate each expert's reference-answer loss and identify the lowest-loss expert as the supervision target. We then jointly train the input encoder and a two-layer MLP to predict these targets from request content. All quantized experts remain frozen throughout training. Reference answers are used only for offline supervision; online routing depends solely on the request input.

\subsubsection{Router Architecture}
\label{sec:router-architecture}

The router consists of a Qwen3-Embedding-0.6B~\cite{zhang2025qwen3} input encoder and a two-layer MLP. Initialized from pretrained parameters, the encoder produces contextual representations of the request text. We use the hidden state of the last non-padding token as the request representation and apply L2 normalization to obtain a 1024-dimensional semantic vector. This vector is passed directly to the MLP for expert prediction.

The MLP comprises two fully connected layers. The semantic vector first passes through LayerNorm, followed by the first fully connected layer, a GELU activation, and dropout. The second fully connected layer produces $K$ expert logits, which are converted into expert-selection probabilities through a softmax function.

\subsubsection{Routing Supervision}
\label{sec:routing-supervision}

We construct routing labels from each quantized expert's prediction loss on reference answers. For a training sample $(x,y)$, where $x$ is the request input and $y=(y_1,\ldots,y_T)$ is the reference answer, we define expert $k$'s loss as the mean token-level cross-entropy, $\ell_k(x,y)=-\frac{1}{T}\sum_{t=1}^{T}\log P_k(y_t\mid x,y_{<t})$. Here, $P_k$ denotes expert $k$'s conditional prediction distribution, and $T$ is the number of valid reference-answer tokens. All experts use the same input format and reference answer. Prompt and padding tokens are excluded from the loss calculation to ensure consistent comparisons across experts.

We assign the lowest-loss expert in the candidate set as the supervision label, $z(x,y)\in\arg\min_{k\in\{1,\ldots,K\}}\ell_k(x,y)$, where $K$ is the number of candidate experts. Each sample thus yields an input--expert pair $(x,z)$. Reference answers are used only to determine labels offline; the router itself receives only the request input. All quantized experts remain frozen, allowing their losses to be precomputed and reused when constructing labels for different expert combinations.

This supervision target corresponds to sample-wise minimum-loss selection within the given expert set. For a dataset $\mathcal{D}=\{(x_i,y_i)\}_{i=1}^{N}$, define the total loss of fixed expert $k$ as $L_k=\sum_{i=1}^{N}\ell_k(x_i,y_i)$ and the total loss under oracle routing as $L_{\mathrm{oracle}}=\sum_{i=1}^{N}\min_k\ell_k(x_i,y_i)$. These quantities satisfy the following relationship.

\begin{theorem}[Loss bound under optimal routing]
\label{thm:optimal-routing}
Under oracle routing, $L_{\mathrm{PrismServe}}=L_{\mathrm{oracle}}$. For any fixed expert $k_{\mathrm{fixed}}$,
\begin{equation}
    L_{\mathrm{PrismServe}}
    \leq L_{\mathrm{Single}}
    := L_{k_{\mathrm{fixed}}}.
    \label{eq:oracle-loss-bound}
\end{equation}
Equality holds if and only if
\begin{equation}
    k_{\mathrm{fixed}}\in\arg\min_k\ell_k(x_i,y_i),
    \qquad \forall i\in\{1,\ldots,N\}.
    \label{eq:oracle-equality-condition}
\end{equation}
\end{theorem}

Theorem~\ref{thm:optimal-routing} shows that selecting the lowest-loss expert for each sample yields a total loss no greater than that of any fixed expert; the proof is provided in appendix. PrismServe learns to approximate this selection from request inputs. Its practical gains depend on expert complementarity and routing accuracy, and it does not directly inherit the oracle guarantee. Reference-answer loss serves as a proxy for generation quality, whose improvement is assessed through downstream evaluation.

\subsubsection{Training Objective and Optimization}
\label{sec:router-optimization}

We formulate expert selection as a $K$-class classification task. Given input $x_i$, the router produces an expert-selection distribution $p_{\theta}(k\mid x_i)$ and is supervised by the offline label $z_i=z(x_i,y_i)$. We minimize cross-entropy over expert classes:
\begin{equation}
    \mathcal{L}_{\mathrm{router}}
    =
    -\frac{1}{B}
    \sum_{i=1}^{B}
    \log p_{\theta}(z_i \mid x_i),
    \label{eq:router-training-loss}
\end{equation}
where $B$ is the batch size and $\theta$ comprises the trainable parameters of the input encoder and MLP. This objective increases the probability assigned to the target expert. Gradients propagate through the MLP into the encoder, jointly adapting the request representation and classification head to expert selection. Thus, token-level cross-entropy evaluates experts to construct supervision labels, while expert-classification cross-entropy optimizes the router.

We use AdamW~\cite{DBLP:conf/iclr/LoshchilovH19} with separate parameter groups for the encoder and MLP. The encoder uses a lower learning rate to limit updates to its pretrained representations, while the MLP uses a higher learning rate to learn the mapping from semantic representations to expert choices. Training uses linear learning-rate warmup, cosine decay, and gradient clipping. Learning rates, batch size, training epochs, and other hyperparameters are detailed in appendix.

\subsection{Inference System}
\label{sec:inference_system}

\begin{figure*}[t]
    \centering
    \includegraphics[width=.9\textwidth]{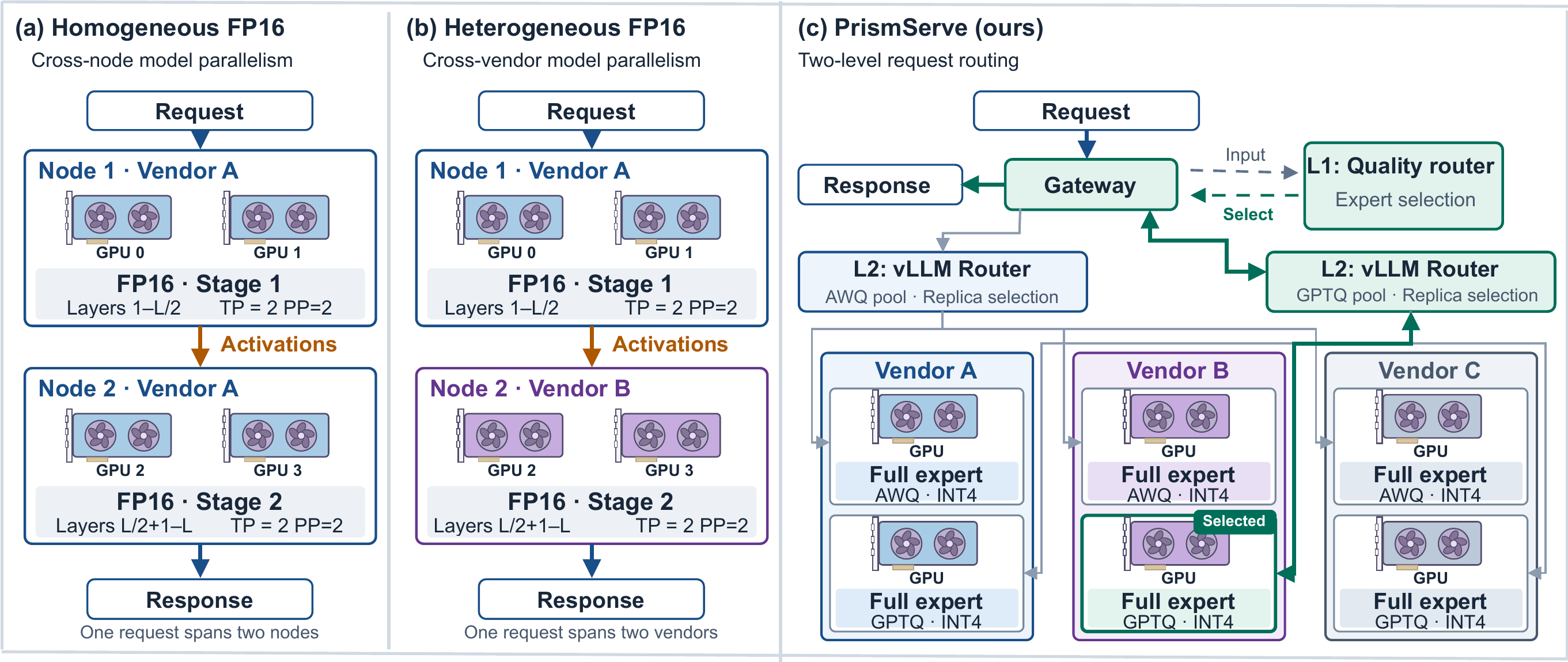}
    \caption{Comparison of inference architectures. (a) Homogeneous FP16 inference partitions a model across nodes from the same hardware vendor. (b) Heterogeneous FP16 inference partitions a model across vendors, requiring cross-vendor activation transfers. (c) PrismServe deploys complete quantized experts across heterogeneous devices. The Gateway obtains an expert-selection decision from the quality router and forwards the request to the corresponding vLLM Router, which selects a replica for complete-model inference.}
    \label{fig:moqe_system}
\end{figure*}
As shown in Figure~\ref{fig:moqe_system} (a)--(c), PrismServe organizes heterogeneous resources through request-level expert routing. FP16 baselines partition a model replica across nodes, requiring intermediate activation transfers between model stages. PrismServe instead deploys complete quantized expert replicas on compatible hardware, allowing heterogeneous nodes to serve different requests independently. Each replica contains all model layers and may use model parallelism within a node.

The system comprises a Gateway, a first-level quality router, second-level replica routers, and expert inference instances. The quality router selects a quantized expert from the request input, while the replica router selects an execution instance within that expert's pool, separating input-dependent quality selection from runtime load balancing. The system builds on the vLLM inference engine~\cite{DBLP:conf/sosp/KwonLZ0ZY0ZS23} and official vLLM Router~\cite{vllmrouter}, with expert selection and request coordination implemented as external services without modifying vLLM source code.

For an automatically routed request, the Gateway sends the input messages to the quality router, which computes the expert-selection probabilities and returns the expert with the highest predicted probability. The Gateway then forwards the original generation request to the corresponding expert pool. The quality router participates only before generation; generation requests and responses pass through the Gateway and downstream serving components. Each pool contains complete replicas of the same quantized model, deployed on hardware compatible with its quantization format and inference backend. Replica counts and placement can change without altering the quality router's output classes.

Each expert pool has a dedicated vLLM Router that selects an available replica according to the configured load-balancing policy. The selected replica performs both prefill and autoregressive decoding, retaining the same expert and execution instance throughout generation. Results are returned through the vLLM Router and Gateway. The green path in Figure~\ref{fig:moqe_system} (c) illustrates a request assigned to the GPTQ pool and subsequently dispatched to a GPTQ INT4 replica on Vendor B.
Each expert pool uses a dedicated vLLM Router with the \texttt{cache\_aware} policy to select an available replica, balancing prefix-cache locality and replica load.
\section{Experiment}
\subsection{Experimental Setup}
\label{sec:experimental-setup}

\textbf{Models and quantized experts.} We evaluate Qwen3-8B, Qwen3-14B, and Qwen3-32B~\cite{DBLP:journals/corr/abs-2505-09388} to examine performance across model scales. For each base model, we construct quantized experts using AWQ~\cite{DBLP:conf/mlsys/0002TTYCWXDG024}, GPTQ~\cite{DBLP:journals/corr/abs-2210-17323}, FlatQuant~\cite{DBLP:conf/icml/SunLBBZLHY0Y0LY25}, QuaRot~\cite{DBLP:conf/nips/AshkboosMCLCJAH24}, and SmoothQuant~\cite{DBLP:conf/icml/XiaoLSWDH23}, with the original full-precision model serving as the quality reference. The quantization configurations are summarized in appendix. All experts derived from the same base model use the same tokenizer, chat template, and context budget. We use Qwen3-14B to evaluate the design, with cross-node and cross-vendor pipeline deployments emulating scenarios where full-precision models exceed single-node memory capacity.

\textbf{Tasks and evaluation metrics.} We evaluate model quality on GSM8K~\cite{DBLP:journals/corr/abs-2110-14168}, MMLU~\cite{hendrycks2020measuring}, IFEval~\cite{zhou2023instruction}, HumanEval~\cite{chen2021evaluating}, and MBPP~\cite{austin2021program}, and assess language modeling performance using perplexity on WikiText-2~\cite{merity2016pointer} and C4~\cite{raffel2020exploring}. We evaluate model quality using the Language Model Evaluation
Harness (lm-eval)~\cite{eval-harness}.

\textbf{Hardware and deployment.} System experiments are conducted on a heterogeneous cluster comprising Huawei Ascend 910C NPUs, Iluvatar CoreX BI-V150 GPUs, and Moore Threads MTT S5000 GPUs. Nodes communicate over the cluster network. Expert instances execute inference using vLLM and the corresponding hardware backends, while the Gateway provides a unified request interface. We compare same-vendor cross-node FP16 inference, cross-vendor cross-node FP16 inference, and PrismServe's deployment of complete quantized expert replicas.

\textbf{Workloads and performance measurement.} We evaluate serving performance using ShareGPT-based requests and synthetic workloads with controlled input and output lengths. All methods use the same request set, workload sequence, and random seeds. After warmup, each configuration is evaluated in three independent runs, each lasting three hours. For each run, output token throughput is computed as the total number of generated output tokens divided by the measurement duration, while TTFT and TPOT percentiles are computed over request-level samples. 

\subsection{Model Quality}
\label{sec:model-quality}

\begin{table*}[t]
    \centering
    \begingroup
    \scriptsize
    \setlength{\tabcolsep}{2pt}
    \setlength{\aboverulesep}{0pt}
    \setlength{\belowrulesep}{0pt}
    \setlength{\extrarowheight}{1pt}
    \renewcommand{\arraystretch}{1.08}
    \resizebox{\textwidth}{!}{%
    \begin{tabular}{|l|*{7}{r}|*{7}{r}|*{7}{r}|}
        \toprule
        \rowcolor{headergray}
        Method & \multicolumn{7}{c|}{Qwen3-8B} & \multicolumn{7}{c|}{Qwen3-14B} & \multicolumn{7}{c|}{Qwen3-32B} \\
        \rowcolor{headergray}
        & GSM & MM & IF & HE & MB & W2 & C4
        & GSM & MM & IF & HE & MB & W2 & C4
        & GSM & MM & IF & HE & MB & W2 & C4 \\
        \midrule
        FP16
        & 88.9 & 74.7 & 84.8 & 87.2 & 63.5 & 9.7 & 13.3
        & 91.4 & 78.5 & 86.3 & 86.6 & 64.3 & 8.6 & 12.0
        & 93.6 & 81.2 & 87.4 & 87.2 & 66.9 & 7.6 & 10.8 \\
        \midrule
        AWQ
        & 83.7 & 67.8 & 77.3 & 79.9 & 56.3 & 12.4 & 16.3
        & 84.9 & 71.8 & 79.3 & 79.3 & 57.1 & 10.8 & 14.6
        & 87.3 & 74.8 & 80.2 & 80.5 & 60.1 & 9.4 & 13.0 \\
        GPTQ
        & 84.3 & 68.9 & 78.2 & 81.1 & 57.4 & 11.5 & 15.4
        & 86.1 & 73.0 & 80.6 & 80.5 & 58.5 & 10.0 & 13.7
        & 88.7 & 76.0 & 81.5 & 81.7 & 61.4 & 8.7 & 12.2 \\
        FlatQuant
        & 85.1 & 70.0 & 79.1 & 81.7 & 58.2 & 10.9 & 14.8
        & 86.8 & 73.8 & 81.5 & 81.1 & 59.0 & 9.5 & 13.2
        & 89.4 & 76.8 & 82.3 & 82.3 & 62.2 & 8.3 & 11.8 \\
        QuaRot
        & 84.5 & 69.3 & 78.0 & 80.5 & 56.9 & 11.2 & 15.1
        & 85.7 & 73.2 & 80.2 & 79.9 & 57.9 & 9.8 & 13.5
        & 88.5 & 76.2 & 81.1 & 81.1 & 60.8 & 8.5 & 12.0 \\
        SmoothQuant
        & 86.1 & 71.5 & 81.0 & 83.5 & 60.1 & 10.5 & 14.3
        & 88.3 & 75.8 & 80.2 & 82.9 & 61.1 & 9.2 & 12.9
        & 89.9 & 77.5 & 84.3 & 79.9 & 61.6 & 8.1 & 11.6 \\
        \midrule
        \textbf{PrismServe}
        & \bestcell{88.4} & \bestcell{74.0} & \bestcell{83.4} & \bestcell{86.0} & \bestcell{62.7} & \bestcell{9.9} & \bestcell{13.5}
        & \bestcell{90.6} & \bestcell{77.8} & \bestcell{85.6} & \bestcell{85.4} & \bestcell{63.2} & \bestcell{8.8} & \bestcell{12.2}
        & \bestcell{93.0} & \bestcell{80.6} & \bestcell{86.5} & \bestcell{86.6} & \bestcell{66.1} & \bestcell{7.8} & \bestcell{11.0} \\
        \bottomrule
    \end{tabular}%
    }
    \endgroup
    \caption{Quality comparison across three Qwen3 model scales. PrismServe uses five experts. GSM: GSM8K; MM: MMLU; IF: IFEval; HE: HumanEval; MB: MBPP; W2: WikiText-2. Higher task scores and lower W2/C4 perplexities are better.}
    \label{tab:model_quality}
\end{table*}

We evaluate whether input-dependent expert selection improves quality over fixed quantized models. Table~\ref{tab:model_quality} compares PrismServe with the FP16 reference and five individual quantization methods across Qwen3-8B, Qwen3-14B, and Qwen3-32B. All PrismServe results use five quantized experts, one each from AWQ, GPTQ, FlatQuant, QuaRot, and SmoothQuant. The router selects one expert per input, whereas each individual-expert baseline uses the same quantized model for all requests. Within each model scale, all methods share the base model, input processing, and evaluation protocols.

PrismServe achieves the highest scores among quantized methods on all five downstream tasks across the three model scales, outperforming the strongest individual expert for each task by 2.0--4.3 points. For example, on Qwen3-14B, PrismServe achieves 85.6 on IFEval, exceeding the strongest individual expert, FlatQuant, by 4.1 points. These improvements hold against the best fixed expert selected separately for each task, supporting the value of input-dependent selection. Across all fifteen model--task combinations, PrismServe reduces the remaining gap to FP16 to 0.5--1.4 points.

The language modeling results show a consistent trend. PrismServe achieves the lowest perplexity among quantized methods on both WikiText-2 and C4 at every model scale. Relative to the strongest individual expert, perplexity decreases by 0.3--0.6 on WikiText-2 and 0.6--0.8 on C4. The remaining gap to FP16 is 0.2 across all six model--dataset combinations. Together, these results show that routing recovers part of the quality lost to quantization across both downstream tasks and language modeling.

PrismServe improves accuracy on Qwen3-14B by 3.6--4.1 points over the best fixed quantized expert, while requiring a single quality-routing step before generation with a mean routing service latency of 47.19\,ms (see Appendices).
\subsection{Effect of Expert Pool Size}
\label{sec:expert-pool-size}
\begin{figure*}[t]
    \centering
    \includegraphics[width=\textwidth]{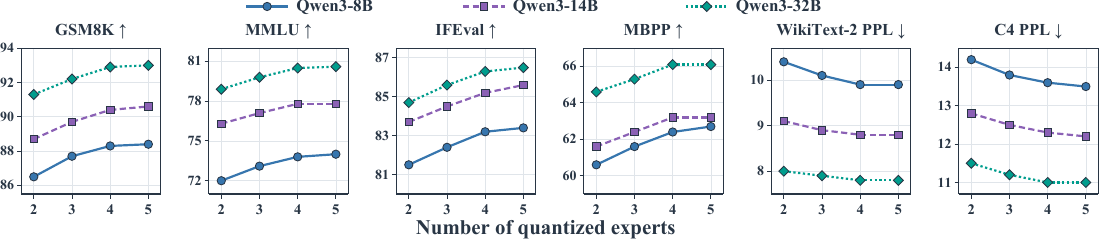}
    \caption{Effect of expert pool size on six benchmarks across three Qwen3 model scales. Starting with GPTQ and AWQ, we sequentially add FlatQuant, QuaRot, and SmoothQuant to form pools of three, four, and five experts. }
    \label{fig:expert_ablation}
\end{figure*}
We examine expert pool size using configurations containing two, three, four, and five quantized experts. Figure~\ref{fig:expert_ablation} reports results separately for GSM8K, MMLU, IFEval, MBPP, WikiText-2, and C4 across the three model scales. The five-expert configuration corresponds to PrismServe in Table~\ref{tab:model_quality}.

Across the evaluated configurations, downstream scores are non-decreasing as the pool expands from two to five experts, improving by 1.5--2.1 points across the twelve model--task combinations. Perplexity also decreases at every model scale, with reductions of 0.2--0.5 on WikiText-2 and 0.5--0.7 on C4. These consistent trends indicate that larger candidate sets provide additional opportunities for input-dependent selection across tasks and model scales.

The marginal gains diminish when expanding from four to five experts. Downstream improvements range from 0 to 0.4 points, WikiText-2 perplexity remains unchanged at all scales, and C4 perplexity decreases by at most 0.1. Thus, four experts capture most of the observed gains for the tested expert compositions, while the fifth provides additional improvements on selected tasks. This result characterizes the evaluated configurations; the benefit of adding an expert also depends on its complementarity with the existing pool.

\subsection{End-to-End Serving Performance}
\label{sec:serving-performance}

All system performance experiments in this section use Qwen3-14B. We compare deployment configurations with 8 and 48 GPUs, reporting output token throughput and P99 TTFT and TPOT.

We construct two workloads. The \texttt{fixed64} workload uses ShareGPT-based requests with input and output lengths fixed at 256 and 64 tokens, respectively. The \texttt{mixed} workload combines synthetic requests with input/output lengths of 256/64, 2,048/64, and 256/1,024 tokens in proportions of 40\%, 30\%, and 30\%, respectively, covering short requests, long inputs, and long outputs. We use five experts for the main quality evaluation and four
for system performance experiments, with the latter's quality
validated on Qwen3-14B in appendix.

\subsubsection{Deployment Comparison with 8 accelerators}
\label{sec:deployment-8cards}
\begin{table*}[t]
    \centering
    \begingroup
    \scriptsize
    \setlength{\tabcolsep}{2.5pt}
    \setlength{\aboverulesep}{0pt}
    \setlength{\belowrulesep}{0pt}
    \setlength{\extrarowheight}{1pt}
    \renewcommand{\arraystretch}{1.12}
    \resizebox{\textwidth}{!}{%
    \begin{tabular}{|ll|c|*{3}{r}|*{3}{r}|*{3}{r}|*{3}{r}|}
        \toprule
        \rowcolor{headergray}
        Deployment & Hardware & TP/PP
        & \multicolumn{3}{c|}{$C=1$}
        & \multicolumn{3}{c|}{$C=2$}
        & \multicolumn{3}{c|}{$C=4$}
        & \multicolumn{3}{c|}{$C=8$} \\
        \rowcolor{headergray}
        & &
        & Thr.$\uparrow$ & TTFT$\downarrow$ & TPOT$\downarrow$
        & Thr.$\uparrow$ & TTFT$\downarrow$ & TPOT$\downarrow$
        & Thr.$\uparrow$ & TTFT$\downarrow$ & TPOT$\downarrow$
        & Thr.$\uparrow$ & TTFT$\downarrow$ & TPOT$\downarrow$ \\
        \midrule
        \multirow{3}{*}{\shortstack[l]{Single-node\\FP16}}
        & Ascend 910C & 2/1
        & 28.64 & \bestcell{56.38} & 36.97
        & 51.32 & \bestcell{109.51} & 40.92
        & 101.34 & \bestcell{164.68} & 41.14
        & 189.34 & \bestcell{267.77} & 43.16 \\
        & MTT S5000 & 2/1
        & \bestcell{39.73} & 84.01 & \bestcell{26.65}
        & 66.55 & 124.04 & 31.69
        & 118.88 & 177.59 & 35.31
        & 222.94 & 314.85 & 36.13 \\
        & BI-V150 & 2/1
        & 32.11 & 99.76 & 33.02
        & 58.77 & 231.32 & 34.70
        & 110.18 & 369.43 & 36.18
        & 205.13 & 635.11 & 35.96 \\
        \midrule
        \multirow{3}{*}{\shortstack[l]{Same-vendor\\cross-node}}
        & Ascend 910C & 2/2
        & 18.12 & 178.57 & 62.23
        & 32.76 & 301.35 & 68.69
        & 60.32 & 424.66 & 73.94
        & 118.58 & 526.38 & 75.11 \\
        & MTT S5000 & 2/2
        & 27.98 & 183.71 & 39.59
        & 52.50 & 338.06 & 41.10
        & 95.89 & 548.37 & 43.64
        & 168.58 & 961.96 & 47.80 \\
        & BI-V150 & 2/2
        & 24.93 & 265.65 & 44.12
        & 34.89 & 440.84 & 63.28
        & 49.76 & 751.13 & 88.31
        & 89.54 & 1616.14 & 92.11 \\
        \midrule
        \multirow{2}{*}{\shortstack[l]{Cross-vendor\\cross-node}}
        & Ascend +  MTT S5000 & 2/2
        & 20.27 & 481.10 & 57.75
        & 34.01 & 729.35 & 67.96
        & 56.70 & 960.86 & 81.89
        & 96.49 & 1335.67 & 95.06 \\
        & MTT S5000 + BI-V150 & 2/2
        & 19.54 & 553.25 & 59.46
        & 29.46 & 675.46 & 79.54
        & 54.11 & 1070.98 & 84.52
        & 91.35 & 1818.33 & 96.89 \\
        \midrule
        \textbf{PrismServe}
        & Heterogeneous & 2/1
        & 36.19 & 132.64 & 29.53
        & \bestcell{68.25} & 277.21 & \bestcell{30.11}
        & \bestcell{127.07} & 419.68 & \bestcell{31.33}
        & \bestcell{226.95} & 834.05 & \bestcell{31.60} \\
        \bottomrule
    \end{tabular}%
    }
    \endgroup
    \caption{8 GPUs deployment comparison under fixed64. TP/PP denotes tensor and pipeline parallelism per replica. Single-node FP16 uses four replicas; cross-node baselines use two. PrismServe uses four quantized experts with one replica each. $C$ denotes total concurrency. Thr. is output throughput in token/s; TTFT and TPOT are P99 latencies in ms and ms/token. }
    \label{tab:deployment_8cards}
\end{table*}
As shown in Table~\ref{tab:deployment_8cards}, we compare nine deployment configurations, each using eight GPUs, under the \texttt{fixed64} workload at total concurrency levels of 1, 2, 4, and 8. Single-node FP16 uses TP=2 and PP=1, with four replicas occupying two GPUs each. Both same-vendor and cross-vendor cross-node FP16 configurations use TP=2 and PP=2, with two replicas occupying four GPUs each. PrismServe deploys four quantized experts, each with one two-GPU replica using TP=2 and PP=1. The hardware composition and parallelism settings are listed in the table.

PrismServe achieves the highest throughput at concurrency levels of 2, 4, and 8, reaching 68.25, 127.07, and 226.95 token/s, respectively. At concurrency 8, its throughput is 1.35--2.53 times that of the same-vendor cross-node configurations and 2.35--2.48 times that of the cross-vendor configurations. It also exceeds the three single-node FP16 configurations by 1.8--19.9\%. These results show that request-level deployment of quantized experts achieves higher aggregate throughput with the same GPU count, particularly relative to cross-node model parallelism.

PrismServe also achieves the lowest TPOT P99 at concurrency levels of 2, 4, and 8. At concurrency 8, its TPOT P99 is 31.60 ms/token, a reduction of 66.8--67.4\% relative to the cross-vendor configurations and 12.1--26.8\% relative to the single-node configurations. As concurrency increases from 1 to 8, its TPOT P99 rises only from 29.53 to 31.60 ms/token, indicating that PrismServe increases aggregate throughput while maintaining comparatively stable decoding tail latency over the evaluated concurrency range.

TTFT reveals a different tradeoff. At concurrency 8, PrismServe's TTFT P99 is 834.05 ms, 37.6--54.1\% lower than the two cross-vendor configurations but higher than the three single-node configurations, which range from 267.77 to 635.11 ms. At concurrency 1, the MTT S5000 single-node configuration also achieves higher throughput than PrismServe, at 39.73 versus 36.19 token/s. PrismServe's main advantages in the eight-GPU experiment therefore lie in throughput at higher concurrency and decoding tail latency. At the same time, single-node deployments retain advantages in TTFT and some low-concurrency settings.

\subsubsection{End-to-End Performance on Accelerator Clusters}
\label{sec:performance-48cards}

We evaluate end-to-end performance on a heterogeneous cluster of 48 accelerators across six nodes, with two nodes each for Ascend 910C, MTT S5000, and BI-V150. Table~\ref{tab:deployment_48cards} compares five configurations: same-node FP16, same-vendor cross-node FP16, cross-vendor cross-node FP16, quantized replicas, and PrismServe. All configurations use the \texttt{fixed64} and \texttt{mixed} workloads at a total request concurrency of 48.

Same-node FP16, quantized replicas, and PrismServe all use TP=2 and PP=1, deploying twenty-four complete-model replicas with two same-type accelerators per replica within a node and four replicas per node. The quantized-replica baseline serves only AWQ models. PrismServe uses four experts---AWQ, GPTQ, FlatQuant, and QuaRot---with six replicas per expert, using two-level routing to select an expert and then an execution replica.

Both cross-node FP16 configurations use TP=2 and PP=2, deploying twelve replicas with four accelerators per replica. Each replica's two pipeline stages reside on separate nodes from the same vendor or different vendors, respectively. In the same-vendor configuration, each vendor's two nodes jointly host four replicas. All configurations confine tensor parallelism to same-type accelerators within a single node.

Under \texttt{fixed64}, PrismServe achieves 1,231.80 token/s, corresponding to 1.12, 2.03, and 2.50 times the throughput of the same-node, same-vendor cross-node, and cross-vendor cross-node FP16 baselines, respectively. Relative to these baselines, P99 TTFT decreases by 23.6\%, 49.3\%, and 67.8\%, while P99 TPOT decreases by 34.5\%, 55.3\%, and 65.4\%. Under \texttt{mixed}, PrismServe achieves 1,365.13 token/s, corresponding to 1.11, 2.08, and 2.63 times the throughput of the three baselines in the same order. P99 TTFT decreases by 10.3\%, 34.7\%, and 63.1\%, while P99 TPOT decreases by 38.1\%, 55.1\%, and 65.4\%. These results show performance improvements over all evaluated FP16 deployments, including the same-node baseline with the same replica count and parallelism configuration.

The quantized-replica baseline achieves the highest throughput and lowest tail latencies on both workloads. Compared with this baseline, PrismServe's throughput is lower by 1.89\% on \texttt{fixed64} and 0.38\% on \texttt{mixed}, P99 TTFT is higher by 47.19 ms on both workloads, and P99 TPOT is higher by 3.91\% and 2.47\%, respectively. Together with the quality evaluation, these results characterize the trade-off between serving a fixed quantized model and selecting quantized experts based on request content.

\begin{table}[t]
    \centering
    \begingroup
    \small
    \setlength{\tabcolsep}{3pt}
    \setlength{\aboverulesep}{0pt}
    \setlength{\belowrulesep}{0pt}
    \setlength{\extrarowheight}{0pt}
    \renewcommand{\arraystretch}{0.95}
    \begin{tabular}{|l|r|r|r|}
        \toprule
        \rowcolor{headergray}
        Deployment & Thr.$\uparrow$ & TTFT$\downarrow$ & TPOT$\downarrow$ \\
        \midrule
        \multicolumn{4}{|l|}{\textbf{fixed64}} \\
        FP16 same-vendor  & 606.79  & 2053.71 & 77.43 \\
        FP16 cross-vendor & 492.47  & 3235.74 & 99.96 \\
        FP16 same-node    & 1102.70 & 1364.97 & 52.76 \\
        Quantized replicas
        & \bestcell{1255.56} & \bestcell{995.09} & \bestcell{33.28} \\
        \textbf{PrismServe}
        & 1231.80 & 1042.28 & 34.58 \\
        \midrule
        \multicolumn{4}{|l|}{\textbf{mixed}} \\
        FP16 same-vendor  & 656.53  & 4874.83 & 80.50 \\
        FP16 cross-vendor & 518.79  & 8638.47 & 104.44 \\
        FP16 same-node    & 1233.98 & 3548.68 & 58.34 \\
        Quantized replicas
        & \bestcell{1370.35} & \bestcell{3137.20} & \bestcell{35.27} \\
        \textbf{PrismServe}
        & 1365.13 & 3184.39 & 36.14 \\
        \bottomrule
    \end{tabular}
    \endgroup
    \caption{Performance with 48 GPUs across six nodes
    (two per vendor) at concurrency 48.
    The quantized-replica baseline serves FlatQuant models,
    with each replica executing entirely within one node.
    Throughput: token/s; P99 TTFT: ms; P99 TPOT: ms/token.}
    \label{tab:deployment_48cards}
\end{table}

\subsection{Serving Performance under SLO Constraints}
\label{sec:slo-goodput}
Unlike the closed-loop experiments in \S\ref{sec:serving-performance}, which maintain a fixed concurrency of 48 requests, this experiment uses an open-loop workload with specified request arrival rates. Requests are issued independently of previous completions, allowing the number of outstanding requests to grow subject to the admission limits.

We evaluate PrismServe's effective serving capacity on 48 accelerators under different request arrival rates, comparing it with same-vendor and cross-vendor cross-node full-precision pipelined deployments. We use a \texttt{mixed} workload and set arrival rates to 18, 24, 32, 48, and 64 requests/s. The per-request TPOT SLO threshold is 100 ms/token. The admission layer allows up to 512 in-flight requests and an additional queue of 128 waiting requests, with a maximum queueing time of 10 seconds. The client-side request timeout is 150 seconds, including queueing, routing, and generation. All deployments use the same request workload and admission limits.

We measure goodput as the rate of successfully completed requests that satisfy the TPOT SLO, in requests/s. Rejected, timed-out, failed, and SLO-violating requests do not contribute to goodput. This metric captures both successful request processing and compliance with the decoding-latency requirement.

\begin{table}[t]
    \centering
    \begingroup
    \small
    \setlength{\tabcolsep}{3pt}
    \setlength{\aboverulesep}{0pt}
    \setlength{\belowrulesep}{0pt}
    \setlength{\extrarowheight}{0pt}
    \renewcommand{\arraystretch}{0.95}
    \begin{tabular}{|c|r|r|r|}
        \toprule
        \rowcolor{headergray}
        \shortstack{Arrival rate\\(req/s)}
        & \multicolumn{3}{c|}{Goodput$\uparrow$ (req/s)} \\
        \midrule
        \rowcolor{headergray}
        & \textbf{PrismServe}
        & Same-vendor
        & Cross-vendor \\
        \midrule
        18 & \bestcell{18.00} & 4.29 & 1.02 \\
        24 & \bestcell{18.80} & 4.17 & 0.87 \\
        32 & \bestcell{17.37} & 4.09 & 0.95 \\
        48 & \bestcell{16.59} & 3.96 & 0.96 \\
        64 & \bestcell{16.44} & 3.86 & 1.00 \\
        \bottomrule
    \end{tabular}
    \endgroup
    \caption{Goodput on 48 accelerators under the
    \texttt{mixed} workload at different request arrival rates.
    The TPOT SLO threshold is 100 ms/token.}
    \label{tab:slo-goodput}
\end{table}

As shown in Table~\ref{tab:slo-goodput}, PrismServe achieves the highest goodput at every evaluated arrival rate, delivering 4.19--4.51$\times$ that of the same-vendor cross-node baseline and 16.44--21.61$\times$ that of the cross-vendor cross-node baseline. At 18 requests/s, PrismServe achieves a goodput of 18 requests/s. Its goodput peaks at 18.80 requests/s when the arrival rate reaches 24 requests/s. In contrast, the two baselines achieve only 4.29 and 1.02 requests/s at the lowest evaluated arrival rate, indicating that their ability to serve requests within the TPOT constraint is already limited at this load.

As the arrival rate increases further, PrismServe's goodput gradually decreases to 16.44 requests/s at 64 requests/s, approximately 12.6\% below its peak. Increasing the offered load therefore does not necessarily increase useful service: requests may be rejected, time out, or complete without satisfying the TPOT requirement. Nevertheless, PrismServe retains a substantial advantage at the highest evaluated load. These results demonstrate higher SLO-constrained serving capacity than the two full-precision cross-node pipelined deployments under the same accelerator budget and admission constraints.

\subsection{Scalability}
\label{sec:scalability}

We evaluate PrismServe's scalability using 48, 64, 80, and 88 accelerators. Starting from the 48-accelerator cluster, we add Ascend and Iluvatar accelerators while keeping the number of Moore Threads accelerators unchanged. All configurations use the mixed workload and the same SLO threshold, admission limits, and request timeout as in \S\ref{sec:slo-goodput}. We vary the request arrival rate from 18 to 64 requests/s to examine how additional resources affect goodput under different loads.

\begin{figure*}[t]
    \centering
    \includegraphics[width=.9\textwidth]{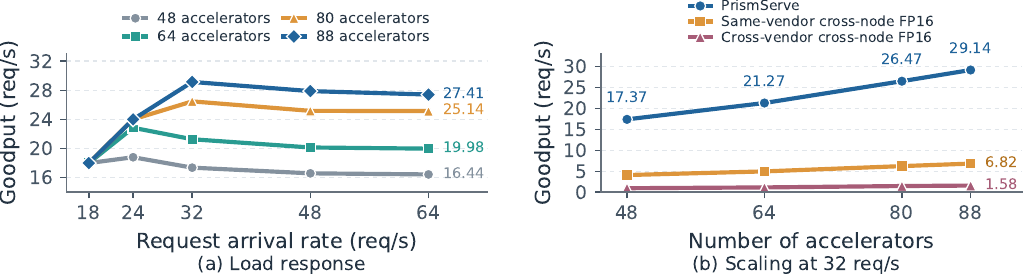}
    \caption{Scalability under the mixed workload. (a) PrismServe's load response at different accelerator counts. (b) Goodput of the three deployments as the accelerator count increases, with the arrival rate fixed at 32 requests/s. The TPOT SLO threshold is 100 ms/token.}
    \label{fig:scalability}
\end{figure*}

Figure~\ref{fig:scalability}(a) shows that additional accelerators improve PrismServe's serving capacity as the offered load increases. At 18 requests/s, all configurations achieve a goodput of 18 requests/s, so the request supply limits the observable scaling benefit. At higher arrival rates, the configurations separate: peak goodput reaches 18.80, 22.85, 26.47, and 29.14 requests/s with 48, 64, 80, and 88 accelerators, respectively. Both the 80- and 88-accelerator configurations sustain a goodput of 24 requests/s at an arrival rate of 24 requests/s. Beyond their respective peaks, goodput declines as the arrival rate increases, indicating that higher offered load does not necessarily translate into more SLO-compliant completions. Nevertheless, at 64 requests/s, increasing the accelerator count from 48 to 88 raises goodput from 16.44 to 27.41 requests/s.

Figure~\ref{fig:scalability}(b) compares all three deployments at a fixed arrival rate of 32 requests/s. PrismServe's goodput increases from 17.37 requests/s with 48 accelerators to 29.14 requests/s with 88 accelerators, a gain of 67.8\%. Over the same range, the same-vendor cross-node baseline increases from 4.09 to 6.82 requests/s, while the cross-vendor cross-node baseline increases from 0.95 to 1.58 requests/s. At 88 accelerators, PrismServe delivers 4.27$\times$ and 18.44$\times$ their goodput, respectively.

\section{Related Work}
\label{sec:related-work}

\textbf{LLM quantization.} Existing methods reduce memory and execution costs through fine-grained quantization, outlier handling, and calibration~\cite{DBLP:journals/corr/abs-2208-07339,yao2022zeroquant,shao2024omniquant}. Weight compression methods further improve low-bit accuracy through sparse representations, sensitivity-aware quantization, and additive codebooks~\cite{dettmers2023spqr,kim2023squeezellm,egiazarian2024extreme}, while incoherence processing and transformation-based approaches reshape numerical distributions to facilitate quantization~\cite{chee2023quip,tseng2024quip,DBLP:conf/nips/AshkboosMCLCJAH24,lin2024duquant,liu2025spinquant}. 

\textbf{Input-adaptive model routing.} Model cascades and quality predictors select models based on request characteristics or intermediate responses to balance quality and cost~\cite{chen2023frugalgpt,aggarwal2024automix,ding2024hybrid}. RouteLLM, RouterDC, and MixLLM learn input--model compatibility through preference supervision, contrastive learning, and contextual bandits, respectively~\cite{ong2025routellm,chen2024routerdc,wang2025mixllm}. RouterBench and LLMRouterBench provide unified evaluations of routing strategies~\cite{hu2024routerbench,li2026llmrouterbench}. 

\textbf{Efficient and heterogeneous serving.} Existing systems improve serving efficiency through scheduling, KV-cache management, prefix reuse, and request migration~\cite{yu2022orca,DBLP:conf/sosp/KwonLZ0ZY0ZS23,agrawal2024taming,zheng2024sglang,sun2024llumnix}. DistServe and Splitwise separate prefill and decoding to enable independent resource allocation~\cite{zhong2024distserve,patel2024splitwise}, while HexGen, HexGen-2, and Helix optimize model placement and execution across heterogeneous devices~\cite{jiang2023hexgen,jiang2025hexgen,mei2025helix,jiang2026boute,DBLP:conf/ppopp/ZhaoWWPL24}. 

\section{Conclusion}
\label{sec:conclusion}
We present PrismServe, which deploys fully quantized experts within homogeneous accelerator groups and uses two-level routing to select experts and replicas, avoiding cross-type communication during generation. Across three Qwen3 model sizes, it improves accuracy over the best fixed quantized expert by 2.0--4.3 points while remaining within 1.4 points of full precision. On 48 accelerators, PrismServe achieves 2.0--2.6$\times$ the throughput of FP16 cross-node deployments and reduces P99 TPOT by 55--65\%, with throughput within 1.9\% of fixed quantized replicas. For a fixed expert pool, replica counts and placement can change without router retraining. Scaling to 88 accelerators further increases SLO-constrained goodput, demonstrating the value of quantization diversity and request-level routing for heterogeneous serving.

\bibliography{example_paper}
\bibliographystyle{mlsys2025}
\section*{AI Usage Disclosure}
We used ChatGPT to assist with language polishing and proofreading of the manuscript. The authors take full responsibility for the accuracy and integrity of the final paper.


\appendix
\clearpage
\section{Experimental Environment and Deployment Configurations}
\label{app:deployment}

This appendix supplements the hardware environment, software stack, model-parallel configurations, and replica placement used in the system performance experiments.

\subsection{Hardware and Software Environment}
\label{app:hardware-software}

Our experiments use three hardware platforms: Huawei Ascend 910C, Moore Threads MTT S5000, and Iluvatar CoreX BI-V150.

\begin{table}[h]
    \centering
    \begingroup
    \small
    \setlength{\tabcolsep}{5pt}
    \setlength{\aboverulesep}{0pt}
    \setlength{\belowrulesep}{0pt}
    \renewcommand{\arraystretch}{1.12}
    \begin{tabular}{|l|c|c|}
        \toprule
        \rowcolor{headergray}
        Hardware & GPUs per node & Memory \\
        \midrule
        Ascend 910C & 8 & 128 GB \\
        MTT S5000 & 8 & 80 GB \\
        BI-V150 & 8 & 32 GiB \\
        \bottomrule
    \end{tabular}
    \endgroup
    \caption{Hardware configurations. Memory is reported per board for Ascend 910C and MTT S5000, and per runtime-visible logical device for BI-V150.}
    \label{tab:app-hardware}
\end{table}

Each platform uses a compatible PyTorch distribution, vendor extension or runtime, and vLLM backend. Table~\ref{tab:app-software} lists the software environments of representative nodes from the three platforms.

\begin{table}[h]
    \centering
    \begingroup
    \scriptsize
    \setlength{\tabcolsep}{3pt}
    \setlength{\aboverulesep}{0pt}
    \setlength{\belowrulesep}{0pt}
    \renewcommand{\arraystretch}{1.12}
    \resizebox{\columnwidth}{!}{%
    \begin{tabular}{|l|l|l|l|}
        \toprule
        \rowcolor{headergray}
        Node & PyTorch & Extension/runtime & vLLM \\
        \midrule
        \texttt{ascend910c-01}
        & 2.7.1
        & torch-npu 2.7.1
        & 0.11.0 \\
        \texttt{s5000-01}
        & 2.5.0
        & torch-musa 2.5.0
        & 0.9.3.dev0 \\
        \texttt{iluvatar150-01}
        & 2.10.0+CoreX 4.5
        & CoreX 4.5
        & 0.23.0 \\
        \bottomrule
    \end{tabular}%
    }
    \endgroup
    \caption{Software environments of representative nodes.}
    \label{tab:app-software}
\end{table}

\subsection{Model Parallelism and Replica Accounting}
\label{app:replica-accounting}

TP and PP denote the tensor-parallel and pipeline-parallel degrees within a model replica, respectively. A complete replica contains all model layers and can independently serve a generation request. Pipeline stages within a replica are not counted as separate replicas.

All configurations use the same GPU count as their resource budget and allocate replicas according to model precision and execution strategy. The FP16 cross-node baselines use TP=2 and PP=2, requiring four GPUs per replica. PrismServe uses quantized experts with TP=2 and PP=1, requiring two GPUs per replica. This configuration allows PrismServe to deploy more complete expert replicas within the same resource budget.

\subsection{Deployment with Eight GPUs}
\label{app:deployment-8gpus}

The eight-GPU experiment compares nine configurations: three single-node FP16 configurations, three same-vendor cross-node FP16 configurations, two cross-vendor cross-node FP16 configurations, and PrismServe. All configurations are evaluated under the \texttt{fixed64} workload at total concurrency levels of 1, 2, 4, and 8.

\textbf{Single-node FP16.} Each hardware platform uses TP=2 and PP=1 to deploy four replicas within a single node, with two GPUs per replica. Each request is assigned to one replica, which performs both prefill and autoregressive decoding.

\textbf{Same-vendor cross-node FP16.} Each hardware platform uses TP=2 and PP=2 to deploy two replicas. The two pipeline stages of each replica reside on two nodes from the same vendor, with two GPUs performing tensor parallelism within each stage.

\textbf{Cross-vendor cross-node FP16.} We evaluate Ascend--Moore and Moore--Iluvatar configurations, both using TP=2 and PP=2 with two replicas. The two GPUs within each stage belong to the same vendor, while the two pipeline stages reside on nodes from different vendors.

\textbf{Replica load balancing.} Each expert pool uses the \texttt{cache\_aware} policy provided by the official vLLM Router. The policy favors prefix-cache locality when replica loads are balanced and prioritizes less-loaded replicas when the configured imbalance thresholds are exceeded. Requests are distributed only among replicas of the expert selected by the quality router, preserving the expert identity.

\textbf{PrismServe.} We deploy four quantized experts with one replica per expert. Each replica uses TP=2 and PP=1, occupies two GPUs, and executes the complete quantized model on compatible hardware. The four replicas collectively use eight GPUs.

\subsection{Deployment with Forty-Eight GPUs}
\label{app:deployment-48gpus}

The forty-eight-GPU experiment compares same-vendor cross-node FP16, cross-vendor cross-node FP16, and PrismServe. All configurations use forty-eight GPUs and are evaluated at total concurrency 48 under both the \texttt{fixed64} and \texttt{mixed} workloads.

\textbf{Same-vendor cross-node FP16.} The cluster is divided into three vendor-specific groups, each containing two nodes and sixteen GPUs. Each replica uses TP=2 and PP=2, with its two pipeline stages placed on the two nodes within the group. Each group hosts four replicas, yielding twelve replicas across the cluster. Here, ‘’same-vendor'' describes the scope of model parallelism within each replica; the three vendor groups collectively provide serving capacity by processing different requests.

\textbf{Cross-vendor cross-node FP16.} This configuration also uses TP=2 and PP=2 with twelve replicas. The two pipeline stages of each replica reside on nodes from different vendors, while tensor parallelism within each stage uses GPUs from the same vendor. Consequently, each request requires intermediate activation transfers across vendors.

\textbf{PrismServe.} We deploy four quantized experts---AWQ, GPTQ, FlatQuant, and QuaRot---with six replicas per expert, yielding twenty-four replicas. Each replica uses TP=2 and PP=1 on two GPUs within a single node, and each node hosts four replicas. Model-parallel communication remains within individual nodes, while heterogeneous nodes contribute serving capacity by processing different requests.

\subsection{Configuration Summary}
\label{app:configuration-summary}

Table~\ref{tab:app-deployment-summary} summarizes the parallelism, replica counts, and placement strategies used in the two deployment scales.

\begin{table*}[t]
    \centering
    \begingroup
    \small
    \setlength{\tabcolsep}{4pt}
    \setlength{\aboverulesep}{0pt}
    \setlength{\belowrulesep}{0pt}
    \renewcommand{\arraystretch}{1.12}
    \resizebox{\textwidth}{!}{%
    \begin{tabular}{|c|l|c|c|c|c|l|}
        \toprule
        \rowcolor{headergray}
        Total GPUs & Deployment & TP & PP & GPUs/replica & Replicas & Placement \\
        \midrule
        8 & Single-node FP16
        & 2 & 1 & 2 & 4
        & Four replicas within one node \\
        8 & Same-vendor cross-node FP16
        & 2 & 2 & 4 & 2
        & Each replica spans two same-vendor nodes \\
        8 & Cross-vendor cross-node FP16
        & 2 & 2 & 4 & 2
        & Each replica spans nodes from different vendors \\
        8 & PrismServe
        & 2 & 1 & 2 & 4
        & Four experts, one replica per expert \\
        \midrule
        48 & Same-vendor cross-node FP16
        & 2 & 2 & 4 & 12
        & Four replicas per vendor group \\
        48 & Cross-vendor cross-node FP16
        & 2 & 2 & 4 & 12
        & Pipeline stages placed across vendors \\
        48 & PrismServe
        & 2 & 1 & 2 & 24
        & Six replicas per expert; four replicas per node \\
        \bottomrule
    \end{tabular}%
    }
    \endgroup
    \caption{Deployment configurations for the eight-GPU and forty-eight-GPU experiments. TP and PP are specified per replica.}
    \label{tab:app-deployment-summary}
\end{table*}

PrismServe organizes replicas into pools by quantized expert. The first-level quality router determines the expert identity, and the second-level instance router selects an execution replica within that expert's pool. The selected replica performs both prefill and decoding, with the expert identity and execution replica remaining unchanged throughout generation.

\subsection{Scaling Configurations}
\label{app:scaling-config}

Table~\ref{tab:scaling-config} summarizes the hardware
configurations used in the scalability experiments.
The initial 48-accelerator cluster contains two nodes
from each vendor. We expand it to 64 accelerators by
adding eight Ascend 910C and eight BI-V150 accelerators,
and repeat the same expansion to reach 80 accelerators.
Finally, we add eight Ascend 910C accelerators to reach 88.
The number of MTT S5000 accelerators remains unchanged.
Each node contains eight accelerators, and each expert
replica uses two same-type accelerators within a single
node with TP=2 and PP=1.

\begin{table}[ht]
    \centering
    \begingroup
    \scriptsize
    \setlength{\tabcolsep}{3pt}
    \renewcommand{\arraystretch}{1.05}
    \begin{tabular}{r|rrr|rr}
        \toprule
        \rowcolor{headergray}
        Total & Ascend & Moore & Iluvatar & Nodes & Replicas \\
        \midrule
        48 & 16 & 16 & 16 & 6  & 24 \\
        64 & 24 & 16 & 24 & 8  & 32 \\
        80 & 32 & 16 & 32 & 10 & 40 \\
        88 & 40 & 16 & 32 & 11 & 44 \\
        \bottomrule
    \end{tabular}
    \endgroup
    \caption{Scaling configurations. Ascend, Moore, and
    Iluvatar denote the numbers of Ascend 910C, MTT S5000,
    and BI-V150 accelerators, respectively.
    Replicas denote the total number of PrismServe
    expert replicas.}
    \label{tab:scaling-config}
\end{table}
\section{Quantized Experts and Router Training Details}
\label{app:router-training}

This appendix supplements the quantized expert configurations, the sources and composition of the router training data, and the router architecture and optimization settings.

\subsection{Quantized Expert Configurations}
\label{app:quantized-experts}
\begin{table}[t]
    \centering
    \begingroup
    \small
    \setlength{\tabcolsep}{4pt}
    \setlength{\aboverulesep}{0pt}
    \setlength{\belowrulesep}{0pt}
    \renewcommand{\arraystretch}{1.1}
    \begin{tabular}{|p{0.46\columnwidth}|p{0.44\columnwidth}|}
        \toprule
        \rowcolor{headergray}
        Quantized expert & Configuration \\
        \midrule
        AWQ         & W4A16 \\
        GPTQ        & W4A16 \\
        FlatQuant   & W4A16 \\
        QuaRot      & W4A16 \\
        SmoothQuant & W8A8 \\
        \midrule
        \rowcolor{headergray}
        Calibration parameter & Setting \\
        \midrule
        Corpus & C4 \\
        Number of samples & 256 \\
        Maximum sequence length & 512 tokens \\
        Random seed & 42 \\
        \bottomrule
    \end{tabular}
    \endgroup
    \caption{Expert precision configurations and shared calibration settings. W and A denote weight and activation bit widths, respectively.}
    \label{tab:app-quantization}
\end{table}
For Qwen3-8B, Qwen3-14B, and Qwen3-32B, we construct five quantized experts using AWQ, GPTQ, FlatQuant, QuaRot, and SmoothQuant. Experts derived from the same base model share the tokenizer, chat template, and context budget to ensure consistent input processing.

AWQ, GPTQ, FlatQuant, and QuaRot use W4A16 quantization with a weight group size of 128, whereas SmoothQuant uses W8A8. All methods use the same calibration samples from C4. Table~\ref{tab:app-quantization} summarizes the precision configurations and calibration settings.

\subsection{Router Training Data and Supervision}
\label{app:router-data}

The router training data comprises problem-solving, conversation, and document samples, accounting for 80\%, 10\%, and 10\% of the samples, respectively. Problem-solving samples cover mathematical reasoning, knowledge-based question answering, and code generation. Conversation samples provide interaction requests with dialogue context, while document samples provide language modeling examples. Table~\ref{tab:app-router-data} lists the data sources.

\begin{table}[H]
    \centering
    \begingroup
    \small
    \setlength{\tabcolsep}{3pt}
    \setlength{\aboverulesep}{0pt}
    \setlength{\belowrulesep}{0pt}
    \renewcommand{\arraystretch}{1.12}
    \begin{tabular}{|p{0.27\columnwidth}|c|p{0.43\columnwidth}|}
        \toprule
        \rowcolor{headergray}
        Category & Proportion & Data sources \\
        \midrule
        Problem-solving
        & 80\%
        & GSM8K, MATH, and MBPP training sets; MMLU auxiliary training set \\
        \midrule
        Conversation
        & 10\%
        & ShareGPT90K \\
        \midrule
        Document
        & 10\%
        & WikiText-2 and C4 training splits \\
        \bottomrule
    \end{tabular}
    \endgroup
    \caption{Sources and composition of the router training data. Proportions are measured by sample count.}
    \label{tab:app-router-data}
\end{table}

All samples are represented as input--reference pairs $(x,y)$. For problem-solving samples, the reference target is the corresponding answer or code. For conversation samples, the input contains the retained dialogue context, and the reference target is the corresponding assistant response. For document samples, the input and reference target correspond to a text prefix and its continuation, respectively.

For each sample, we compute every quantized expert's mean target-token negative log-likelihood using the same input and reference target. The loss covers only reference-target tokens, excluding prompt and padding tokens. The expert with the lowest loss within the current candidate set provides the routing supervision label. Reference targets are used only to construct offline supervision.

We construct labels from the respective quantized experts of Qwen3-8B, Qwen3-14B, and Qwen3-32B and independently train a router for each model scale. When the number or composition of experts changes, we reconstruct the labels within the current candidate set and retrain the router. Quantized experts remain frozen, allowing their precomputed losses to be reused across different expert subsets.

\subsection{Router Architecture and Optimization}
\label{app:router-optimization}
\begin{table}[t]
    \centering
    \begingroup
    \footnotesize
    \setlength{\tabcolsep}{4pt}
    \setlength{\aboverulesep}{0pt}
    \setlength{\belowrulesep}{0pt}
    \renewcommand{\arraystretch}{1.08}
    \begin{tabular}{|p{0.44\columnwidth}|p{0.46\columnwidth}|}
        \toprule
        \rowcolor{headergray}
        Parameter & Setting \\
        \midrule
        Input encoder
        & Qwen3-Embedding-0.6B \\
        Training strategy
        & Joint encoder and MLP optimization \\
        Quantized expert parameters
        & Frozen \\
        Pooling
        & Last non-padding token \\
        Semantic vector normalization
        & L2 normalization \\
        Semantic vector dimension
        & 1024 \\
        MLP input dimension
        & 1024 \\
        MLP hidden dimension
        & 64 \\
        MLP output dimension
        & Number of experts $K$ \\
        Activation
        & GELU \\
        Dropout
        & 0.1 \\
        Training objective
        & Cross-entropy over expert labels \\
        Optimizer
        & AdamW \\
        Encoder learning rate
        & $10^{-6}$ \\
        MLP learning rate
        & $10^{-3}$ \\
        Weight decay
        & 0.01 \\
        Batch size
        & 40 \\
        Epochs
        & 8 \\
        Warmup ratio
        & 5\% \\
        Learning-rate decay
        & Cosine decay \\
        Minimum learning-rate ratio
        & 10\% of initial learning rate \\
        Gradient clipping threshold
        & 1.0 \\
        Training seeds
        & 2026, 42, 7 \\
        Maximum encoder input length
        & 32,768 tokens \\
        \bottomrule
    \end{tabular}
    \endgroup
    \caption{Router architecture and training hyperparameters.}
    \label{tab:app-router-hyperparameters}
\end{table}
The router comprises a Qwen3-Embedding-0.6B input encoder and a two-layer MLP. We use the hidden state of the last non-padding token as the request representation and apply L2 normalization to obtain a 1024-dimensional semantic vector. 

The classifier consists of LayerNorm, a fully connected layer, GELU activation, dropout, and an output layer. Its hidden dimension is 64, and its output dimension equals the number of candidate experts $K$. A softmax function converts the output logits into expert-selection probabilities.

We jointly optimize the input encoder and MLP using cross-entropy loss over expert labels, while keeping all quantized experts frozen. We use AdamW with learning rates of $10^{-6}$ for the encoder and $10^{-3}$ for the MLP. The first 5\% of training steps use linear warmup, followed by cosine decay to 10\% of each parameter group's initial learning rate. Table~\ref{tab:app-router-hyperparameters} summarizes the configuration.

\subsection{Evaluation Protocol}
\label{app:evaluation-protocol}

We use the Language Model Evaluation Harness (lm-eval) for downstream task evaluation. GSM8K uses 8-shot chain-of-thought prompting, MMLU uses 5-shot prompting, IFEval and HumanEval use zero-shot evaluation, and MBPP uses 3-shot prompting. We evaluate language modeling performance using token-level perplexity (PPL) on WikiText-2 and C4. Table~\ref{tab:app-evaluation-protocol} summarizes the evaluation settings and metrics.

\begin{table}[t]
    \centering
    \begingroup
    \small
    \setlength{\tabcolsep}{4pt}
    \setlength{\aboverulesep}{0pt}
    \setlength{\belowrulesep}{0pt}
    \renewcommand{\arraystretch}{1.1}
    \begin{tabular}{|l|c|l|}
        \toprule
        \rowcolor{headergray}
        Benchmark & Few-shot & Metric \\
        \midrule
        GSM8K      & 8-shot CoT & Accuracy \\
        MMLU       & 5-shot     & Accuracy \\
        IFEval     & 0-shot     & Instruction-following rate \\
        HumanEval  & 0-shot     & pass@1 \\
        MBPP       & 3-shot     & pass@1 \\
        WikiText-2 & --         & Token-level PPL \\
        C4         & --         & Token-level PPL \\
        \bottomrule
    \end{tabular}
    \endgroup
    \caption{Evaluation settings and metrics. CoT denotes chain-of-thought prompting.}
    \label{tab:app-evaluation-protocol}
\end{table}

\section{Expert Selection Analysis}
\label{app:expert-selection}

We analyze the fraction of requests assigned to each quantized expert across three model scales and expert pools containing two, three, and four experts. Selection share is computed as the number of requests assigned to an expert divided by the total number of requests. We use the \texttt{mixed} synthetic workload, combining input/output lengths of 256/64, 2,048/64, and 256/1,024 tokens in proportions of 40\%, 30\%, and 30\%, respectively. The two-expert pool contains GPTQ and AWQ; the three- and four-expert pools additionally include FlatQuant and QuaRot, respectively. Each configuration uses its corresponding trained router.

\begin{figure*}[t]
    \centering
    \includegraphics[width=\textwidth]{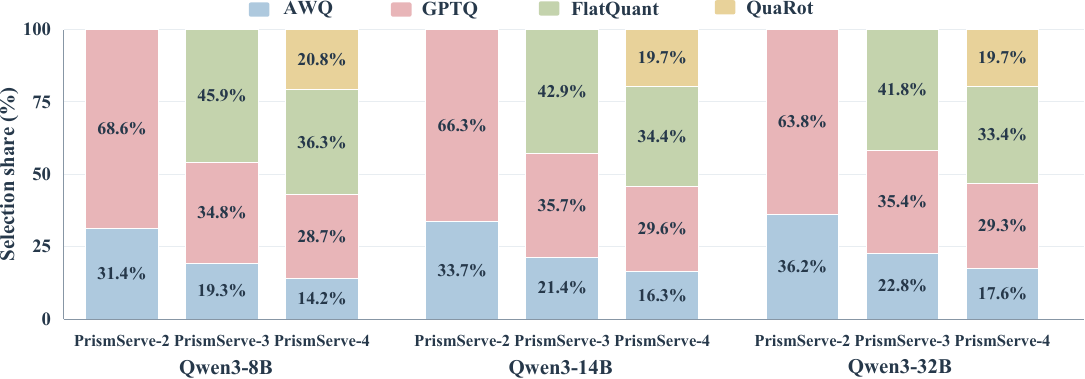}
    \caption{Expert selection shares under the \texttt{mixed} synthetic workload. Each model group contains PrismServe configurations with two, three, and four experts. Each bar sums to 100\%, and segment labels indicate the percentage of requests assigned to each expert.}
    \label{fig:expert-selection}
\end{figure*}

As shown in Figure~\ref{fig:expert-selection}, expert selection is nonuniform and exhibits similar patterns across model scales. In the two-expert configuration, GPTQ receives 63.8--68.6\% of requests. With FlatQuant included, the three-expert configuration assigns the largest share to FlatQuant, ranging from 41.8\% to 45.9\%. Thus, the most frequently selected expert changes with the candidate pool.

In the four-expert configuration, FlatQuant remains the most frequently selected expert, receiving 33.4--36.3\% of requests, while QuaRot receives 19.7--20.8\%. All three model scales exhibit the same selection order: FlatQuant, GPTQ, QuaRot, and AWQ. Nevertheless, every candidate expert receives a nonzero share, with individual shares ranging from 14.2\% to 36.3\%. The router therefore uses multiple experts under this workload rather than consistently assigning requests to a single expert.

\section{Routing Effectiveness and Overhead}
\label{app:routing}

\subsection{Routing Effectiveness}
\label{app:routing-effectiveness}

We evaluate the effectiveness of input-dependent expert selection on Qwen3-14B using a four-expert pool comprising AWQ, GPTQ, FlatQuant, and QuaRot. We compare three routing strategies: PrismServe-4, uniform random routing, and reverse selection. PrismServe-4 assigns each request to the expert with the highest predicted probability; uniform random routing selects each expert with a probability of $25\%$; reverse selection uses the same router but assigns each request to the expert with the lowest predicted probability. All three strategies execute only the selected expert and use identical test sets, input processing, and task evaluation settings. We also report results for the full-precision model and individual quantization methods. SmoothQuant serves only as an additional fixed-expert reference and is excluded from the four-expert pool.

\begin{table*}[t]
    \centering
    \begingroup
    \small
    \setlength{\tabcolsep}{5pt}
    \setlength{\aboverulesep}{0pt}
    \setlength{\belowrulesep}{0pt}
    \renewcommand{\arraystretch}{1.12}
    \begin{tabular}{|l|c|c|c|c|c|}
        \toprule
        \rowcolor{headergray}
        Method & GSM8K $\uparrow$ & MMLU $\uparrow$ & IFEval $\uparrow$ & HumanEval $\uparrow$ & MBPP $\uparrow$ \\
        \midrule
        FP16 & \textbf{91.4} & \textbf{78.5} & \textbf{86.3} & \textbf{86.6} & \textbf{64.3} \\
        \midrule
        AWQ & 84.9 & 71.8 & 79.3 & 79.3 & 57.1 \\
        GPTQ & 86.1 & 73.0 & 80.6 & 80.5 & 58.5 \\
        FlatQuant & 86.8 & 73.8 & 81.5 & 81.1 & 59.0 \\
        QuaRot & 85.7 & 73.2 & 80.2 & 79.9 & 57.9 \\
        SmoothQuant & 88.3 & 75.8 & 80.2 & 82.9 & 61.1 \\
        \midrule
        Uniform random routing & 85.9 & 73.0 & 80.4 & 80.2 & 58.1 \\
        Reverse selection & 82.4 & 69.1 & 76.3 & 76.5 & 54.2 \\
        \textbf{PrismServe-4} & \textbf{90.4} & \textbf{77.8} & \textbf{85.2} & \textbf{85.4} & \textbf{63.2} \\
        \bottomrule
    \end{tabular}
    \endgroup
    \caption{Quality comparison of the full-precision model, individual quantized experts, and routing strategies on Qwen3-14B. All three routing strategies use the same four-expert pool: AWQ, GPTQ, FlatQuant, and QuaRot. SmoothQuant is excluded from routing and serves only as an additional fixed-expert reference. All scores are percentages; higher is better.}
    \label{tab:app-routing-effectiveness}
\end{table*}

As shown in Table~\ref{tab:app-routing-effectiveness}, PrismServe-4 outperforms uniform random routing on all five tasks by $4.5$--$5.2$ points. Since both strategies use the same expert pool and execute only one expert per request, these results support the effectiveness of input-dependent expert selection: the gains cannot be explained solely by randomly mixing quantized variants. Reverse selection performs worst on every task, trailing PrismServe-4 by $8.0$--$9.0$ points. This further supports the router's ability to distinguish experts that are more or less suitable for a given request.

Compared with the best fixed expert within the four-expert pool on each task, PrismServe-4 improves quality by $3.6$--$4.3$ points. Even when SmoothQuant is included in the comparison, PrismServe-4 exceeds the best individual quantization method by $2.0$--$3.7$ points and remains within $0.7$--$1.2$ points of the full-precision model. These results show that learned routing exploits complementarity among quantized experts to recover most of the quality lost to quantization, supporting the four-expert configuration used in our end-to-end serving experiments.

\subsection{Quality Router Serving Overhead}
\label{app:routing-overhead}

We deploy the quality router on a single BI-V150 accelerator and measure routing service latency for individual requests, including queueing, gateway forwarding, and router computation. Table~\ref{tab:app-routing-overhead} summarizes the results.

\begin{table}[t]
    \centering
    \begingroup
    \small
    \setlength{\tabcolsep}{5pt}
    \setlength{\aboverulesep}{0pt}
    \setlength{\belowrulesep}{0pt}
    \renewcommand{\arraystretch}{1.12}
    \begin{tabular}{|l|c|}
        \toprule
        \rowcolor{headergray}
        Metric & Latency (ms) \\
        \midrule
        Mean latency & 47.19 \\
        P95 latency & 52.90 \\
        P99 latency & 55.94 \\
        \bottomrule
    \end{tabular}
    \endgroup
    \caption{Routing service latency, including queueing, gateway forwarding, and router computation.}
    \label{tab:app-routing-overhead}
\end{table}

The routing service has a mean latency of $47.19$~ms and a P99 latency of $55.94$~ms. Under the 48-accelerator configuration with total concurrency 48 reported in Table~3, the ratio of routing service P99 latency to P99 time to first token (TTFT) is $5.37\%$ for \texttt{fixed64} and $1.76\%$ for \texttt{mixed}. Quality routing is performed once when a request enters the system and is not repeated during autoregressive decoding, so this overhead is incurred before the first token is returned.

\section{Theoretical Analysis and Proof of Theorem~\ref{thm:optimal-routing}}
\label{app:theoretical-analysis}

\subsection{Definition of Loss Upper Bound}

We refer to Theorem~4.1 of to analyze quantization-induced loss. Under their signed symmetric uniform weight-quantization setting and corresponding assumptions, the excess-loss bound for a data subset depends on the quantization step size, the loss gradient norm, and the Hessian trace. Introducing an expert index $k$, we denote this upper-bound expression by
\begin{equation}
\begin{aligned}
U_k(a)\triangleq{}&
\frac{1}{2}\sqrt{n}(s_{\max}^{(k)})^{2}
\left\|g_{w^{*}}^{D_a}\right\|\\
&+\frac{1}{8}n(s_{\max}^{(k)})^{4}
\mathrm{Tr}\!\left(H_{w^{*}}^{D_a}\right)\\
&+\mathcal{O}\!\left(\|\Delta w_k^{*}\|^{3}\right).
\end{aligned}
\label{eq:app-quantization-bound}
\end{equation}

Here, $w^{*}$ denotes the pretrained model parameters, $n$ is the parameter dimension, $\Delta w_k^{*}$ is the weight perturbation introduced by expert $k$, and $s_{\max}^{(k)}$ is its largest quantization step size. The gradient and Hessian are defined as
\begin{equation}
\begin{aligned}
g_{w^{*}}^{D_a}
&=\nabla_w\mathcal{L}(w^{*};D_a),\\
H_{w^{*}}^{D_a}
&=\nabla_w^{2}\mathcal{L}(w^{*};D_a),
\end{aligned}
\end{equation}
where $\mathcal{L}(w;D_a)$ denotes the loss evaluated on subset $D_a$ with parameters $w$.

This result indicates that quantization-induced loss depends on both numerical precision and the gradient and curvature associated with the data subset. Different inputs may therefore exhibit different sensitivities to quantization, motivating input-dependent expert selection. The cited bound concerns excess loss relative to the unquantized model; below, we directly prove Theorem~\ref{thm:optimal-routing} using the reference-answer loss defined in the main text.

\subsection{Proof of Theorem~\ref{thm:optimal-routing}}

\begin{proof}
For any fixed expert $k_{\mathrm{fixed}}$ and any sample $(x_i,y_i)$,
\begin{equation}
\min_k\ell_k(x_i,y_i)
\leq \ell_{k_{\mathrm{fixed}}}(x_i,y_i).
\end{equation}
Summing over all samples and applying the definition of oracle routing gives
\begin{equation}
\begin{aligned}
L_{\mathrm{PrismServe}}
&=L_{\mathrm{oracle}}\\
&=\sum_{i=1}^{N}\min_k\ell_k(x_i,y_i)\\
&\leq\sum_{i=1}^{N}\ell_{k_{\mathrm{fixed}}}(x_i,y_i)\\
&=L_{\mathrm{Single}}.
\end{aligned}
\label{eq:app-oracle-routing-bound}
\end{equation}

Since each sample-level loss gap is nonnegative, equality of the total losses holds if and only if every sample-level inequality holds with equality. Equivalently,
\begin{equation}
k_{\mathrm{fixed}}
\in\arg\min_k\ell_k(x_i,y_i),
\qquad \forall i\in\{1,\ldots,N\},
\end{equation}
allowing ties among experts.
\end{proof}

\subsection{Expert Complementarity and Low-Precision Quantization}

The equality condition in Theorem~\ref{thm:optimal-routing} characterizes when oracle routing provides no loss reduction over a particular fixed expert: that expert must attain the minimum loss on every sample. Conversely, if another expert achieves strictly lower loss on at least one sample, oracle routing achieves strictly lower total loss than that fixed expert. The potential benefit of routing therefore depends on sample-level expert complementarity, rather than differences in average expert quality alone.

The analysis of \citet{DBLP:conf/icml/LiuMZZYWLD25} further identifies gradient norms and Hessian traces as factors affecting quantization sensitivity across data subsets. This motivates examining whether selecting a fixed expert solely by its average quality overlooks other experts' advantages on particular inputs. PrismServe evaluates this complementarity empirically and trains a router to select quantized experts from request inputs.

Under otherwise identical uniform quantization settings with the same weight range, reducing the bit width increases the quantization step size. The scale-dependent terms in Equation~\ref{eq:app-quantization-bound} consequently grow, indicating potentially greater loss sensitivity at lower precision. This motivates studying expert routing in low-precision regimes. However, larger quantization errors alone do not guarantee greater routing benefits; practical gains depend on expert complementarity and the learned router's ability to select suitable experts.
\end{document}